# Capsule Network-Based Semantic Intent Modeling for Human-Computer Interaction

Shixiao Wang
School of Visual Arts
New York, USA

Yifan Zhuang
University of Southern California
Los Angeles, USA

Runsheng Zhang
University of Southern California
Los Angeles, USA

Zhijun Song*
Parsons school of Design
New york, USA

*Abstract-This paper proposes a user semantic intent modeling algorithm based on Capsule Networks to address the problem of insufficient accuracy in intent recognition for human-computer interaction. The method represents semantic features in input text through a vectorized capsule structure. It uses a dynamic routing mechanism to transfer information across multiple capsule layers. This helps capture hierarchical relationships and part-whole structures between semantic entities more effectively. The model uses a convolutional feature extraction module as the low-level encoder. After generating initial semantic capsules, it forms high-level abstract intent representations through an iterative routing process. To further enhance performance, a margin-based mechanism is introduced into the loss function. This improves the model's ability to distinguish between intent classes. Experiments are conducted using a public natural language understanding dataset. Multiple mainstream models are used for comparison. Results show that the proposed model outperforms traditional methods and other deep learning structures in terms of accuracy, F1-score, and intent detection rate. The study also analyzes the effect of the number of dynamic routing iterations on model performance. A convergence curve of the loss function during training is provided. These results verify the stability and effectiveness of the proposed method in semantic modeling. Overall, this study presents a new structured modeling approach to improve intent recognition under complex semantic conditions.*

I. INTRODUCTION

With the rapid development of intelligent technologies, human-computer interaction (HCI) systems are evolving from traditional command-based inputs to more natural, efficient, and intelligent forms[1]. In various application scenarios, such as intelligent customer service, voice assistants, mobile app recommendations, and question answering systems, accurate understanding of user intent has become central to delivering high-quality interactions. The essence of HCI lies in understanding human behavioral motives and expression contexts. Semantic intent serves as the bridge in this process, playing a key role in converting user input into system behavior. Therefore, extracting accurate and deep semantic intent from complex user expressions has become a critical direction in current HCI research[2].

Traditional intent recognition methods rely heavily on shallow feature extraction and classification models. These approaches have limited capacity to model complex semantic structures and contextual variations. When faced with ambiguity, polysemy, and context dependence in user inputs, they often lack robustness and generalization. In natural language [3], word relationships are not simply linear. They exhibit hierarchical and compositional structures [4]. Conventional neural networks struggle to capture these part-whole semantic relationships, which restricts their ability to model complex user intent [5]. To address these challenges, developing models that better simulate human cognitive structures has emerged as a promising solution[6].

Capsule Networks, a recent deep learning architecture, show significant advantages in modeling spatial hierarchies and capturing feature composition patterns. Unlike traditional neurons, capsules represent entities as vectors or matrices [7]. This allows them to retain more information about "pose" and "relations." Through dynamic routing, they can map low-level features to high-level semantic concepts with high precision [8]. This mechanism aligns well with how humans process semantic composition and decomposition in language. Capsule Networks are expected to offer stronger generalization and semantic expression in intent modeling. Introducing Capsule Networks into the intent recognition domain not only complements existing techniques but also explores a new paradigm for semantic modeling[9].

In complex HCI scenarios, user expressions are often diverse, personalized, and ambiguous [10]. Extracting clear and directed semantic intent from such input is essential for accurate system understanding and high interaction quality. Traditional models struggle to capture the dynamic evolution of semantics[11]. In contrast, the structured representation and multidimensional mapping of Capsule Networks enable finer-grained modeling of semantic hierarchies. Especially in challenging contexts such as short texts, non-standard input, and compound intents, Capsule Networks can preserve structural relationships between semantic elements through vector representation [12-14]. This offers new insights for developing intent modeling algorithms with better interpretability and reasoning capacity[15].

This study aims to address real-world HCI needs by leveraging the semantic modeling potential of Capsule Networks to build an intent recognition model with hierarchical semantic expression. By structurally analyzing and reconstructing the latent semantics in user input, the goal is to enhance the system's intelligence and understanding of user intent. This approach holds practical significance for improving system response accuracy and user experience. It also lays a foundation for downstream tasks such as semantic understanding, multi-turn dialogue, and context awareness [16-18]. In the context of evolving semantic modeling technologies, exploring the integration of Capsule Networks into intent recognition is expected to promote theoretical advancement and practical applications in natural language processing and human-computer interaction.

## II. METHOD

This study develops a user semantic intent modeling algorithm grounded in the capsule network framework. Building on the insights of Sun [19], who emphasized adaptive semantic structure optimization in interactive interfaces, this model adopts a vectorized representation to capture semantic entities in user inputs such as commands or sentences. These vectorized forms serve as low-level features that are progressively refined into higher-level abstract representations. At the initial stage, user inputs undergo word vector encoding to construct foundational semantic embeddings. These embeddings are then passed through a convolutional layer, which acts as a low-level encoder to extract local contextual features. Subsequently, leveraging the dynamic routing mechanism characteristic of capsule networks, the model aggregates these features into capsule structures, thereby preserving the hierarchical and compositional relationships among semantic units. The model architecture is presented in Figure 1.

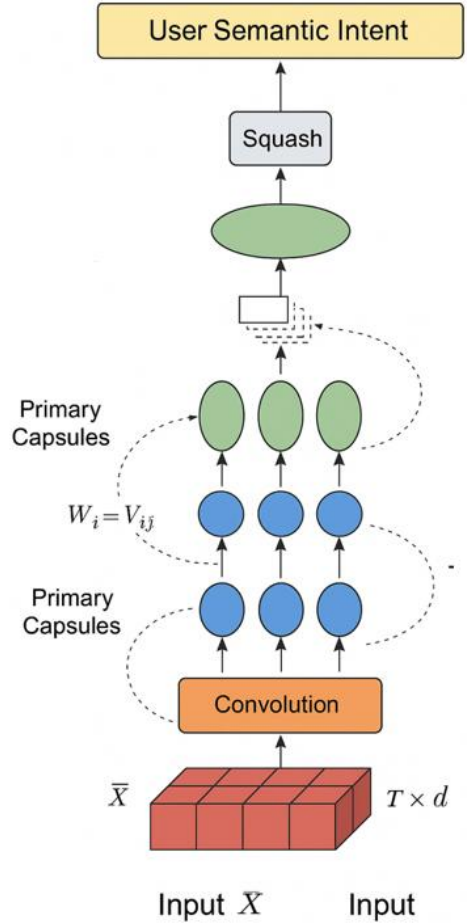

Figure 1. Overall model architecture diagram

Assume that the user input consists of T words, each word is mapped to a d-dimensional word vector, forming an input matrix $X \in R^{T \times d}$, which is expressed as:

$$X = [x_1, x_2, ..., x_T], \ x_i \in R^d$$

Next, the input is subjected to several convolution operations to extract local semantic features, resulting in a set of low-level semantic capsules. Each capsule uses a vector to represent a feature entity. If the l-th layer contains $n_l$ capsules, each capsule is a u-dimensional vector, then this layer is represented by $U^{(l)} \in R^{n_l \times u}$.

A dynamic routing relationship is established between low-level capsules and high-level capsules. The core is that the low-level capsule $u_i$ is projected to the high-level candidate capsule $u_i$ through the weight $c_{ij}$ and the transformation matrix $W_{ij}$. The calculation method is:

$$\hat{u}_{j|i} = W_{ij} u_i$$

The high-level capsule $s_j$ is obtained by weighted summing of all projections from the lower levels:

$$s_j = \sum_i c_{ij} \hat{u}_{j|i}$$

Afterwards, in order to enhance the nonlinear modeling capability, the "squash" nonlinear function is applied to the aggregation result $s_j$ to compress it into the unit vector space, representing the existence probability and direction of the high-level semantic entity:

$$v_j = \frac{\|s_j\|^2}{1+\|s_j\|^2} \cdot \frac{s_j}{\|s_j\|}$$

Finally, each capsule vector $v_j$ in the output layer represents a potential semantic intent category, and its modulus $\|v_j\| \in [0,1]$ is used as the confidence that the intent is activated. The model uses the margin loss function to optimize the intent classification effect. The loss function is as follows:

$$L = \sum_j T_j \cdot \max(0, m^+ - \|v_j\|)^2 + \lambda(1-T_j) \cdot \max(0, \|v_j\| - m^-)^2$$

Among them, $T_j \in \{0,1\}$ indicates whether it is a true label, $m^+, m^-$ is the threshold of positive and negative samples, and $\lambda$ is the balance factor. The entire modeling process achieves hierarchical modeling and recognition of user semantic intent through structured semantic expression and spatial transformation mechanism without relying on sequence assumptions or graph structures.

## III. EXPERIMENTAL RESULTS

### A. Dataset

This study adopts the SNIPS Natural Language Understanding Dataset as the primary resource for modeling user semantic intent. The dataset is widely used in natural language understanding tasks. It is especially suitable for intent recognition and slot filling. It effectively reflects the characteristics of real user input in intelligent voice interaction systems. The dataset covers several common scenarios. It offers good diversity and task representativeness, making it appropriate for modeling and validating the effectiveness of semantic intent recognition models.

The SNIPS dataset contains approximately 14,000 user utterances. These cover seven common intent categories, including music playback, weather queries, restaurant search, and alarm setting. The data consists of frequent task commands found in everyday life. The language is natural, and the content is clear. Each entry includes a user input sentence and its corresponding intent label. Some samples also include annotated slot information, which supports further semantic structure analysis.

The dataset provides high-quality semantic annotations and balanced data distribution. It supports the modeling of deep semantic structures. In tasks involving multi-turn dialogue or context awareness, the SNIPS dataset is often used as a standard benchmark. It is widely applied for evaluating and comparing the performance of intent recognition and semantic understanding models. Using this dataset for modeling helps verify the model's ability to capture real user intent in typical human-computer interaction scenarios.

### B. Experimental Results

In this section, this paper first gives the comparative experimental results of the proposed algorithm and other algorithms, as shown in Table 1.

Table 1. Comparative experimental results

| Method | Accuracy (%) | F1-Score | Intent Detection Rate |
|---|---|---|---|
| BiLSTM+Attention[20] | 91.2 | 90.2 | 90.8 |
| CNN-CRF[21] | 89.7 | 88.8 | 88.9 |
| JointBERT[22] | 93.5 | 92.6 | 93.2 |
| Dynamic Capsule NLU[23] | 94.1 | 93.3 | 93.9 |
| Ours | 95.6 | 94.7 | 95.1 |

The experimental results show that the proposed algorithm outperforms mainstream methods in overall performance on the intent recognition task. In terms of accuracy, the method achieves 95.6%, which is significantly higher than 91.2% for BiLSTM+Attention and 89.7% for CNN-CRF. It also performs better than recent strong models such as JointBERT and Dynamic Capsule NLU. This indicates that introducing Capsule Networks for semantic modeling can effectively improve the ability to identify user intent. The model shows stronger generalization, especially in scenarios with complex expressions or semantic ambiguity.

The comparison of F1-Score further confirms the model's balanced performance between precision and recall. The proposed method reaches an F1-Score of 94.7, which is higher than 92.6 for JointBERT and 93.3 for Dynamic Capsule NLU. This suggests that the model provides more stable classification performance across multiple intent categories. Traditional methods like BiLSTM and CNN are good at extracting local features. However, they are limited in modeling semantic structure and context dependence. This is an area where the dynamic routing mechanism of Capsule Networks shows clear advantages.

From the perspective of Intent Detection Rate, the proposed method achieves the highest coverage for correctly recognized intents, reaching 95.1%. This demonstrates that the model can identify not only common intents but also less frequent or composite ones with high accuracy. In contrast, traditional sequence models or position-based convolutional structures often lack hierarchical semantic modeling capabilities. As a result, they struggle to capture deeper intent features and show slightly lower detection rates.

Taken together, these metrics show that the Capsule Network-based semantic intent modeling method for human-computer interaction has clear advantages. It performs well in expressing semantic entities, modeling hierarchical semantic structures, and maintaining classification stability. The model uses vector representation and dynamic aggregation to address information loss in complex language structures. This provides a more semantically expressive approach to intent recognition.

This paper further gives the impact of the number of dynamic routing iterations on the accuracy of intent recognition, and the experimental results are shown in Figure 2.

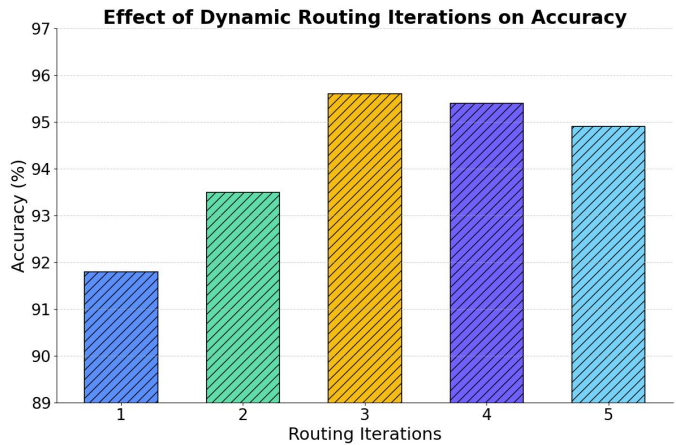

Figure 2. The impact of dynamic routing iterations on intent recognition accuracy

The figure shows that the number of dynamic routing iterations has a significant impact on intent recognition accuracy. The relationship is nonlinear. When the number of iterations is set to 1, the model achieves the lowest accuracy, only 91.8%. This suggests that without sufficient routing updates, lower-level capsules fail to aggregate high-quality high-level semantic representations. As a result, the model's ability to distinguish intents is reduced. At this stage, information transfer is too coarse to build a complete semantic structure.

When the number of iterations increases to 3, accuracy reaches its peak at 95.6%. At this point, the dynamic routing mechanism optimizes the mapping from lower-level capsules to higher-level semantic representations through multiple rounds. The model becomes more precise and stable in capturing user intent. This result indicates that a moderate increase in iterations enhances the capsule network's ability to represent complex semantic structures. It is especially effective for ambiguous, polysemous, or context-rich user input.

However, further increasing the number of iterations to 4 and 5 causes a slight drop in accuracy. This may be due to redundant feature reconstruction introduced by excessive iterations. It can increase the risk of overfitting or reduce the model's semantic generalization. Although dynamic routing improves information selectivity, its marginal benefit decreases after a certain point. It may even interfere with the established aggregation paths.

Therefore, the results suggest that more iterations do not always lead to better performance in intent recognition tasks using capsule networks. There exists an optimal range for the number of routing iterations. Proper control of routing rounds helps model complex semantics while maintaining stability and generalization. This provides a theoretical basis for structural optimization in future human-computer interaction system design.

Finally, the loss function drop graph is given, as shown in Figure 3.

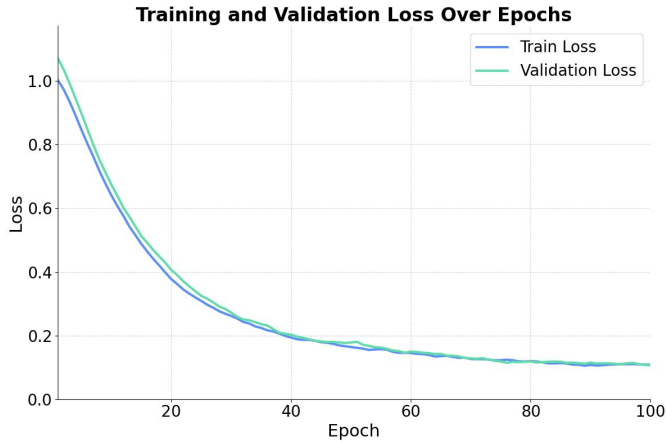

Figure 3. Loss function drop graph

The figure shows that both training loss and validation loss decrease steadily during the training process. After around 30 epochs, they begin to stabilize. This indicates that the proposed model has good convergence behavior during training. In the initial stage, the loss drops rapidly. This suggests that the Capsule Network has strong feature learning ability. It can quickly capture the core semantic information in user input and effectively represent semantic intent.

In addition, the gap between validation loss and training loss remains small. There is no sign of overfitting. This demonstrates that the model has good generalization ability in the user intent modeling task. The collaboration between hierarchical semantic representation and the dynamic routing mechanism enhances this effect. The model not only adapts to the semantic distribution of the training data but also handles unseen semantic variations with stability. This provides a reliable modeling foundation for intent understanding in human-computer interaction systems.

IV. CONCLUSION

This paper addresses the problem of user semantic intent recognition in human-computer interaction. A modeling approach based on Capsule Networks is proposed to overcome the limitations of traditional models in semantic structure representation and hierarchical information capture. By introducing vector-based representation and a dynamic routing mechanism, the method effectively simulates part-whole relationships between semantic entities. This enhances the model's ability to understand complex semantic structures. Experimental results show that the proposed model outperforms mainstream methods across multiple metrics, confirming its effectiveness and superiority in intent recognition tasks.

The study further investigates the impact of iteration times in the dynamic routing mechanism on model performance. It finds that a moderate number of iterations helps optimize the information aggregation process between capsules, thus improving recognition accuracy. This finding provides a theoretical basis for structural tuning in building more efficient and controllable semantic understanding models. In addition, the model shows stable performance in loss convergence, verifying its robustness during both training and generalization. This supports its deployment in real-world interaction systems.

At the application level, the proposed semantic intent modeling method can be widely applied to various typical human-computer interaction scenarios. These include intelligent customer service, voice assistants, mobile app recommendations, and smart dialogue systems. The method's ability to handle fuzzy, unstructured, and short-text input makes it particularly suitable for user input parsing and task-oriented dialogue management in natural language understanding. It can significantly improve the system's ability to interpret user intent and enhance the level of intelligent response, offering strong practical value and broad application potential.

## V. Future Work

Future work may explore multimodal fusion by incorporating prosody, facial expressions, and gestures to enhance intent recognition, as demonstrated in pediatric gait analysis [24]. The integration of large language models (LLMs), particularly those combined with autoencoders and MLPs [25], could improve generalization in intent modeling. In addition, IoT-based sensing enables real-time, scalable deployment and has shown success in medical tasks like skin cancer detection [26]. Finally, broader adoption should consider ethical implications of LLM-driven systems in sensitive domains [27].